\documentclass[11pt]{article}
\usepackage{eacl2014}
\usepackage{times}
\usepackage{url}
\usepackage{latexsym}
\usepackage{amsmath}
\usepackage{attachfile2}
\usepackage{graphicx}
\usepackage{url}

\makeatletter
\def\url@leostyle{%
  \@ifundefined{selectfont}{\def\UrlFont{\sf}}{\def\UrlFont{\small\ttfamily}}}
  \makeatother
\newcommand{\setmetera}[2]{\ensuremath{\genfrac{}{}{0pt}{}{#1}{#2}}} % for time signature

 \title{Generating Music from Literature}

\author{Hannah Davis \\
   New York University\\
   {\tt hannah.davis@nyu.edu} \\\And
   Saif M. Mohammad \\
   National Research Council Canada\\
   {\tt saif.mohammad@nrc-cnrc.gc.ca} \\}

\date{}

\begin{document}
\maketitle
\begin{abstract}
% Generating music is immensely challenging given the infinite possibilities of notes, tempo, melodies, etc.
We present a system, {\it TransProse}, that automatically generates musical pieces from text.
% which capture the use of emotion words in pieces of literary work.
TransProse  uses known relations between elements of music 
such as tempo and scale,
and the emotions they evoke. 
%, such as using major keys to express positive sentiment, 
%and higher tempo to express activity. 
 Further, it uses a novel mechanism to determine sequences of notes
that capture the emotional activity
 in text.
% Audio pieces pertaining to a number of popular novels are embedded in this pdf.
% There are a number of applications for mapping text to music
The work has applications in
information visualization, in creating audio-visual e-books,
and in developing music apps.
% Our system generates
\end{abstract}

\section{Introduction}
Music and literature have an intertwined past. It is believed that they originated together \cite{Brown70},
but in time, the two have developed into separate art forms that continue to influence each other.\footnote{The term {\it music} 
comes from {\it muses}---the nine Greek goddesses of inspiration for literature, science, and arts.}
% Literature has three major genres: prose, drama, and poetry.
Music, just as prose, drama, and poetry, is often used to tell stories.\footnote{Music is especially close to poetry as
songs often tend to be poems set to music.}
Opera and ballet tell stories through music and words,
but even instrumental music, which is devoid of words, can
have a powerful narrative form \cite{Hatten91}. 
Mahler's and Beethoven's symphonies, for example, are regarded as particularly good examples
of narrative and evocative music \cite{Micznik01}.
% Just as good literature, instrumental music is capable of evoking strong emotional responses.

In this paper, for the first time, we present a method to automatically generate music
from literature. Specifically, we focus on novels and generate music
that captures the change in the distribution of emotion words. 
% Giving computers the ability to generate evocative original pieces of music is probably just
% as difficult as making computers generate interesting original stories.
% In the near future at least, we do not expect that computers will generate music 
% as good as that of human composers. 
% Our goal here is simply to produce an audio piece
% that sounds like music (as opposed
% to a disconsonant sequences of notes) and also
% captures, in some small way, the emotional essence of a novel.
% Since there is no previous work that we could find attempting to generate
% music from text, 
We list below some of the 
benefits in pursuing this general line of research:
% Generating music from text has a number of applications, including:
\vspace*{-3mm}
\begin{itemize}
\item Creating audio-visual e-books that generate music when certain pages are opened---music that accentuates the mood conveyed by the text in those pages.
\vspace*{-1mm}
\item Mapping pieces of literature to musical pieces according to compatibility of the flow of emotions in text with the audio characteristics of the musical piece.
\vspace*{-3mm}
\item Finding songs that capture the emotions in different parts of a novel. This could be useful, for example,
 to allow an app to find and play songs that are compatible with the mood of the chapter being read.
\vspace*{-3mm}
\item Generating music for movie scripts.
\vspace*{-3mm}
% \item Consider the scenario where one is listening to music while reading an e-book.
% An app could find and play songs that are compatible to the mood of the chapter being read.
% \vspace*{-3mm}
% \item % Consider the scenario where one is listening to their favorite musical pieces while reading an e-book.
% An app that finds and plays songs that are compatible to the mood of the chapter or book being read.
% \vspace*{-3mm}
\item Appropriate music can add to good visualizations to communicate information
effectively, quickly, and artfully.\\
{\it Example 1}: A tweet stream that is accompanied by
music that captures the aggregated sentiment towards an entity.\\
{\it Example 2}: Displaying the world map where clicking on a particular region plays
music that captures the emotions of the tweets emanating from there.
%  {\it Example 3}: Consider a visualization of novels as per their densities of emotion word usage.
% Clicking on a particular novel
% plays music that evokes the emotions in accordance with the emotion densities. 
\end{itemize}

Given a novel (in an electronically readable form), our system, which we call {\it TransProse}, generates simple piano pieces whose notes
are dependent on the emotion words in the text.
The challenge in composing new music, just as in creating a new story, is the infinite number of choices and
possibilities. 
We present a number of mapping rules to determine various elements of music, such as tempo, major/minor key, etc.\@ according to the emotion 
word density in the text.
We introduce a novel method to determine the sequence of notes (sequences of pitch and duration pairs) to be played 
as per the change in emotion word density in the text. 
We also list some guidelines to make the sequence of notes sound like music as opposed to a cacophonous cascade of sounds.

Certainly, there is no one right way of capturing the emotions in text through music, and there is no one right way to
produce good music. Generating compelling music is an art, and % we do not claim that TransProse is particularly good at it.
% In fact, 
TransProse can be improved in a number of ways
(we list several advancements in the Future Work section). Our goal with this project is to present
initial ideas in an area that has not been explored before.

This paper does not assume any prior knowledge of music theory. Section 2 presents all the terminology and
ideas from music theory needed to understand this paper. We present related work in Section 3. Sections 4, 5, 6,
and 7 describe our system. In Sections 8 and 9, we present an analysis of the music generated by our system 
for various popular novels. Finally in Section 10 we present limitations and future work.

%   -- Talk about how other facets of text may also be reflected in music; but how emotions makes a lot of sense.
%   -- List various challenges in creating such a system.
%   -- Key conclusions. What worked. What did not.

% do not assume any knowledge of music 
\section{Music}
%   -- Music Terminology
%   	-- Describe what basic music terms mean
%		-- the major or minor key, tempo, note length, pitch, melody, and complexity. 
%		-- time signature, tempo, melody, consonant and dissonant intervals
%   	-- Describe what makes good music
%   -- Audio software details
%	-- which pieces of software were used; how can one obtain them; are they free
%	-- what is the input and output of each software used

% For the the purposes of understanding this paper, we present below some fundamentals
% of music. 
% The reader can skip this section if they are already familiar with terms such as
% note, melody, tempo, and octave, and with some basic music theory of consonance. 

In physical terms, music is a series of possibly overlapping sounds, often intended to be pleasing to the listener. 
Sound is what our ears perceive when there is a mechanical oscillation of pressure in some medium such as air or water.
Thus, different sounds are associated with different frequencies.
In music, a particular frequency is referred to as {\it pitch}.
A {\it note} has two aspects: pitch and relative duration.\footnote{Confusingly, {\it note}
is also commonly used to refer to {\it pitch} alone. To avoid misunderstanding, we will not use {\it note}
in that sense in this paper. However, some statements, such as {\it play that pitch} may seem odd to those familiar with music,
who may be more used to {\it play that note}.}
Examples of relative duration are {\it whole-note}, {\it half-note}, {\it quarter-note}, etc.
Each successive element in this list is of half the duration as the preceding element.
Consider the example notes: 400Hz--quarter-note and 760Hz--whole-note. The first note is the sound
corresponding to 400Hz, whereas the second note is the sound corresponding to 760Hz.
Also, the first note is to be played for one fourth the duration the second note is played.
It is worth repeating at this point that {\it note} and {\it whole-note} do not refer to the same concept---the former
is a combination of pitch and relative duration, whereas whole-note (and others such as quarter-note and half-note) are
used to specify the relative duration of a note.
% {\it Rest} is defined as the absence of sound.
% Thus, % if {\it rest} is defined as the absence of sound, 
% music can be defined as a series of notes and rest.
Notes are defined in terms of {\it relative} duration to allow for the same melody to be played quickly or slowly. 
% We will specify how these relative durations are mapped to absolute durations of time shortly.

A series of notes can be grouped into a {\it measure} (also called {\it bar}).
{\it Melody} (also called {\it tune}) is a sequence of measures (and therefore a sequence of notes) that creates the
musical piece itself. For example, a melody could be defined as 620Hz--half-note,
1200Hz-whole-note, 840Hz-half-note, 660Hz--quarter-note, and so on.
There can be one melody (for example, in the song {\it Mary Had A
Little Lamb}) or multiple melodies; they can last throughout the
piece or appear in specific sections.
 A challenge for TransProse is to generate appropriate sequences of notes, given the infinite
 possibilities of pitch, duration, and order of the notes.

{\it Tempo} is the speed at which the piece should be played. It is usually indicated by
the number of beats per minute. % (a beat is analogous to the frequency in a computer CPU). 
A beat is a basic unit of time. A quarter-note is often used as one beat. In which case, the tempo can be understood
simply as the number of quarter-notes per minute. Consider an example. Let's assume it is 
decided that the example melody
specified in the earlier paragraph is to be played at a tempo
of 120 quarter-notes per minute. 
The total number of quarter-notes in the initial sequence (620Hz--half-note,
1200Hz--whole-note, 840Hz--half-note, and 660Hz--quarter-note)
is 2 + 4 + 2 + 1 = 9. Thus the initial sequence must be played in $9/120$ minutes, or $4.5$ seconds.
% For example, some music can have the feel of 1-2-3-4-1-2-3-4. Then such music can be divided into measures with four beats.

The {\it time signature} of a piece indicates two things: a) how many beats are in a measure, and b) which note duration represents one beat. 
It is written as one number stacked on another number. The upper number is the number of beats per measure, 
and the lower number is the note duration that represents one beat. For example, a time signature of \setmetera{6}{8} would mean there are six beats per measure, 
and an eighth note represents one beat. One of the most common time signatures is \setmetera{4}{4}, and it is referred to as {\it common time}.

Sounds associated with frequencies that are multiples or factors of one another (for example, 440Hz, 880Hz, 1760Hz, etc) 
are perceived by the human ear as being consonant and pleasing. This is because the pressure waves associated
with these sounds have overlapping peaks and troughs. 
Sets of such frequencies or pitches form pitch classes. The intervals between successive pitches in a pitch class
are called {\it octaves}. On a modern 88-key piano, the keys are laid out in increasing order of pitch.
Every successive 12 keys pertain to an octave. (Thus there are keys pertaining to 7 octaves and four additional keys pertaining to the eighth octave.) 
Further, each of the 12 keys split the octave such that
the difference in frequency between successive keys in an octave is the same.
Thus the corresponding keys in each octave form a pitch class. For example, the keys at position  1, 
13, 25, 37, and so on, form a pitch class. Similarly keys at position 2, 14, 26, 38, and so on,
form another pitch class. The pitch classes on a piano are given names
C, C\#, D, D\#, E, F, F\#, G, G\#, A, A\#, B. (The \# is pronounced {\it sharp}).
The same names can also be used to refer to a particular key in an octave.
(In an octave, there exists only one C, only one D\#, and so on.)
The octaves are often referred to by a number. On a standard piano, the octaves in increasing order are 0, 1, 2, and so on.
C2 refers to the key in octave 2 that is in the pitch class C.\footnote{The frequencies of piano keys at a given position across octaves is in log scale. For example,
frequencies of C1, C2,\ldots, and so on are in log scale. The perception of sound (frequency) in the human ear is also roughly logarithmic.
Also, the frequency 440Hz (mentioned above) is A4 and it is the customary tuning standard for musical pitch.}

% The difference in frequency between two pitches is called an {\it interval}.
% As mentioned above, the differences in frequencies produced by successive piano keys within an octave
% are equal. 
% As per convention, 
The difference in frequency between successive piano keys is called a {\it semitone} or {\it Half-Tone} ({\it Half} for short).
The interval
between two keys separated by exactly one key is called {\it Whole-Tone} ({\it Whole} for short).
Thus, the interval between C and C\# is half, whereas the interval between C and D is whole.
A {\it scale} is any sequence of pitches ordered by frequency.
A {\it major  scale} is a sequence of pitches obtained by applying the ascending pattern:
Whole--Whole--Half--Whole--Whole--Whole--Half. For example, if one starts with C, then the corresponding
C major scale consists of C, D (frequency of C + Whole interval), E (frequency of D + Whole interval),
F (frequency of E + Half interval), G, A, B, C.
 Major scales can begin with any pitch (not just C), and that pitch is called the {\it base pitch}.
A {\it major key} is the set of pitches corresponding to the major scale. Playing in the key of C major
means that one is primarily playing the keys (pitches) from the corresponding scale, C major scale (although not necessarily in
a particular order).

{\it Minor scales} are series of pitches obtained by applying the ascending pattern:
 Whole-Half--Whole--Whole--Half--Whole--Whole. Thus, C minor is C, D, D\#, F, G, G\#, A\#, C.
%  (It is convention to write minor scales in small letter. For example, c minor or d minor.)
A {\it minor key} is the set of pitches corresponding to the minor scale. 
% Playing the C Minor key means that one is primarily playing the keys from the corresponding scale, C Minor scale.
 Playing in major keys generally creates
 lighter sounding pieces, whereas playing in minor keys creates darker sounding pieces.
% C minor has had the meaning of heroic struggle from Beethoven's time. One of the most famous pieces in this key is his Symphony No. 5. The fact that Brahms's Symphony No. 1 is also in C minor helped it get its nickname as "Beethoven's Tenth" (Beethoven's actual unfinished Symphony No. 10 in E flat major may have had a significant central C minor section in the first movement). Three of Anton Bruckner's ten numbered symphonies are in C minor.

% \item {\it Time Signature}: the rhythm of the piece. The time signature is written as a
% fraction where the numerator is the number of beats per measure, and the
% denominator is the beat as a division of the whole note. For example, a time
% signature of 3/4 would mean that the beat is valued as a quarter note, and there
% are three beats to a measure; a time signature of 3/8 would mean the beat is
% valued as an eighth note, and there are three beats to a measure.

{\it Consonance} is how pleasant or stable
one perceives two pitches played simultaneously (or one after the other).
There are many theories on what makes two pitches consonant, some of which are culturally dependent.
The most
common notion (attributed to Pythagoras) is that the simpler the ratio between the two frequencies, the more consonant 
they are \cite{Roederer2008,Tenney1988}.
% Consonant pitches feel “resolved,” whereas
% dissonant chords feel “unresolved,” meaning the listener feels the
% need for the pitch to move on to a stable or consonant pitch; with consonant
% pitches, the desire for such movement does not exist. 
% Which pitches are
% consonant and which dissonant is somewhat of a historically and culturally
% subjective determination. 
% The order in which intervals are more or less stable
% is not universally agreed upon.
% we use the order proposed in \cite{Lots2008}.

Given a particular scale, some have argued that the order of the pitches in decreasing consonance is as follows:
1st, 5th, 3rd, 6th, 2nd, 4th, and 7th \cite{Perricone00}.
Thus for the C major---C (the base pitch, or 1st), D (2nd) , E (3rd), F (4th), G (5th) , A (6th) , B (7th)---the order of the pitches in decreasing
consonance is---C, G, E, A, D, F, B.
Similarly, for C minor---C (the base pitch, or 1st), D (2nd), D\# (3rd), F (4th), G (5th), G\# (6th), A\# (7th)---the order of pitches in decreasing
consonance is---C, G, D\#, G\#, D, F, A\#.
We will use these orders in TransProse to generate more discordant and unstable pitches to reflect higher emotion word densities in the novels.

\section{Related Work}

This work is related to automatic sentiment and emotion analysis of text (computational linguistics), the generation of music (music theory),
as well as the perception of music (psychology). 
% A comprehensive coverage of all related work is beyond the scope of this paper,
% but we briefly summarize some of the most relevant work.

Sentiment analysis techniques aim to determine the evaluative nature of text---{\it positive, negative,} or {\it neutral}.
They have been applied to many different kinds of texts including customer reviews \cite{PangL08},
newspaper headlines \cite{Bellegarda10}, emails \cite{LiuLS03,MohammadY11},
% blogs \cite{GenereuxE06,MihalceaL06}, 
and tweets \cite{Pak10,Agarwal11,Thelwall11,Brody11,Aisopos12,Bakliwal12}.
Surveys by Pang and Lee \shortcite{PangL08} and Liu and Zhang \shortcite{Liu12} give a summary of many of these approaches.
Emotion analysis and affective computing involve the detection of emotions such as {\it joy, anger, sadness,} and {\it anticipation} in text. A number of approaches for emotion analysis have been proposed in recent years 
\cite{Boucouvalas02,ZheB02,AmanS07,NeviarouskayaPI09,KimGEG2009,Bollen2009,TumasjanSSW2010}.
% HolzmanP03,MaPI05,JohnBX06,MihalceaL06,GenereuxE06,AmanS07,NeviarouskayaPI09,KimGEG2009,Bollen2009,TumasjanSSW2010}.
Text-to-speech synthesis employs emotion detection to produce speech consistent with the emotions in the text \cite{iida00,pierre03,schroder09}.
See surveys by Picard \shortcite{picard00} and Tao and Tan \shortcite{tao05} for a broader review of the research in this area.

Some prior empirical sentiment analysis work focuses specifically on literary texts.
Alm and Sproat \shortcite{Alm05} analyzed 
twenty two Brothers Grimm fairy tales to show that fairy tales often began with a neutral sentence and
ended with a happy sentence.
Mohammad \shortcite{Mohammad12DSS} visualized the emotion word densities in novels and fairy tales.
% Some emotion analysis systems have been evaluated on novels \cite{Boucouvalas02,JohnBX06,FranciscoG06}.
Volkova et al.\@ \shortcite{Volkova10} study human annotation of emotions in fairy tales.
However, there exists no work connecting automatic detection of sentiment with the automatic generation of music.

Methods for both sentiment and emotion analysis often rely on lexicons of words associated with various affect categories such as
positive and negative sentiment, and emotions such as joy, sadness, fear, and anger.
The WordNet Affect Lexicon (WAL) \cite{StrapparavaV04} has a few hundred words
annotated with associations to a number of affect categories including the six Ekman emotions (joy, sadness, anger, fear,
disgust, and surprise).\footnote{{http://wndomains.fbk.eu/wnaffect.html}}
% by manually identifying the emotions of a few seed words and then
% marking all their WordNet synonyms as having the same emotion.
% The words in WAL are annotated for a number of emotion and affect categories, but its creators
% also provided a subset corresponding to the six Ekman emotions.
% General Inquirer (GI) \cite{Stone66} has 11,788 words labeled with 182
% categories of word tags, including positive and negative polarity.\footnote{GI: http://www.wjh.harvard.edu/$\sim$inquirer}
% It has certain other affect categories too, such as arousal and pain, but not categories corresponding to the basic emotions proposed by Ekman. 
% Affective Norms for English Words (ANEW) has pleasure (happy--unhappy), arousal (excited--calm), and dominance (controlled--in control) ratings
% for 1034 words.\footnote{{http://csea.phhp.ufl.edu/media/anewmessage.html}}
The NRC Emotion Lexicon, compiled by Mohammad and Turney \shortcite{MohammadT10,MohammadT13}, has annotations
for about 14000 words with eight emotions (six of Ekman, trust, and anticipation).\footnote{{http://www.purl.org/net/NRCemotionlexicon}}
We use this lexicon in our project.

Automatic or semi-automatic generation of music through computer algorithms was first popularized by Brian Eno (who coined the term {\it generative music}) and David Cope \cite{cope1996experiments}.
Lerdahl and Jackendoff \shortcite{lerdahl1983generative} authored a seminal book on the generative theory of music.
Their work %, as well as work by Eno and Cope,
greatly influenced future work in automatic generation of music such as that of Collins \shortcite{collins2008analysis} and Biles \shortcite{Biles94}.
However, these pieces did not attempt to explicitly capture emotions.

% -- music as a narrative force\\
% -- mapping rules of story to music\\
Dowling and Harwood \shortcite{Dowling86} showed that vast amounts of information are processed when listening
to music, and that the most expressive quality that one perceives is emotion.
The communication of emotions in non-verbal utterances and in music show how emotions in music have an evolutionary basis \cite{rousseau2009essay,spencer1857origin,juslin03}.
There are many known associations between music
and emotions:
\vspace*{-1mm}
\begin{itemize}
\item {\it Loudness}: Loud music is associated with intensity, power, and anger, whereas soft music is associated with sadness or fear \cite{Gabrielsson01}.
% \vspace*{-2mm}
\item {\it Melody}: A sequence of consonant notes is associated with joy and calm, whereas a sequence of disconsonant notes 
is associated with excitement, anger, or unpleasantness \cite{Gabrielsson01}.
\vspace*{-2mm}
\item {\it Major and Minor Keys}: Major keys are associated with happiness, whereas minor keys are associated with sadness \cite{Hunter10,Hunter08,Ali10,Gabrielsson01,Webster05}.
\vspace*{-2mm}
\item {\it Tempo}: Fast tempo is associated with happiness or excitement \cite{Hunter10,Hunter08,Ali10,Gabrielsson01,Webster05}.
\end{itemize}
\vspace*{-2mm}
\noindent % We use these ideas in our system.
Studies have shown that even though many of the associations mentioned above are largely universal, one's own culture
also influences the perception of music \cite{morrison2009cultural,balkwill1999cross}.

% -- Research involving manipulating music or analyzing audio signals for search etc: Shazam like applications

\section{Our System: TransProse}
%    Overview of the system.
%   -- Text analysis
%   -- Mapping textual features to music
%	-- relationship between various affectual features and music 
%		-- may be there should be a table summarizing the chosen mappings
%   	-- what else had to be done to make it musical

% HOW THE ALGORITHM WORKS:

Our system, which we call TransProse, generates music according to the use of emotion
words in a given novel.
It does so in three steps: First, it analyzes the input text and generates an emotion profile.
The emotion profile is simply a collection of various statistics
about the presence of emotion words in the text. Second, based on the emotion profile of the text,
the system generates values for tempo, scale, octave, notes, and the sequence of notes for multiple melodies. 
Finally, these values are provided to JFugue, an open-source Java API for programming music, that generates the appropriate audio file.
In the sections ahead, we describe the three steps in more detail.

\section{Calculating Emotion Word Densities}
 Given a novel in electronic form,  
% the system uses simple heuristics to identify 
% the start of each chapter. For example, the
% presence of chapter headers such as "Chapter" followed by a number or roman numerals.
% If the system is unable to find explicit chapter markers, then it identifies chapter 
% boundaries by the presence of five or more new line characters.
% The system can also analyze a text by simply breaking it into $k$ equal sections, where
% $k$ is is specified by the user.
we use the NRC Emotion Lexicon \cite{MohammadT10,MohammadT13} to identify the number of
words in each chapter that are associated with an affect category.
We generate counts for eight emotions (anticipation, anger, joy, fear, disgust, sadness, 
surprise, and trust) as well as
for positive and negative sentiment.
We partition the novel into four sections
representing the beginning, early
middle, late middle, and end. Each section is further partitioned into four sub-sections.
The number of sections, the number of subsections per section, and the number of notes generated for each of the subsections
together determine the total number of notes generated for the novel. 
Even though we set the number of sections and number of sub-sections to four each,
these settings can be varied, especially for significantly longer or shorter pieces of text.

For each section and for each sub-section the ratio of emotion words to the total number of words
is calculated. We will refer to this ratio as the {\it overall emotions density}. We also calculate
densities of particular emotions, for example,
the {\it joy density}, {\it anger density}, etc.
As described in the section ahead, the emotion densities are used to generate sequences of notes for each of the subsections.
% The number of sections, the number of subsections per section, and the number of notes in the subsections
% together determine the total number of notes generated for the novel. 
% For our current experiments we decided to set the number of sections and number of sub-sections fixed to four each.
% However, these settings can be varied, especially for significantly longer or shorter pieces of text.
% Finally, the affect density scores are linearly interpolated between chapters to create
% a continuous function. As described in the next subsection, this continuous function will be sampled at various points to generate
% appropriate notes, thereby creating the musical piece.

% Using those data points of emotion frequency per chapter, an abstract
% “emotion spline” is created, which connects those points to create an interpolation of
% emotion frequencies that can be sampled at any division of the novel, as opposed to being
% restricted to the specific data points of the individual chapters. This makes it easier to
% experiment with music creation.  

\section{Generating Music Specifications}

% TransProse generates music based on parameters from the emotion profile of the novel,
% and some pre-decided fixed settings.
Each of the pieces presented in this paper are for the piano with three simultaneous, but different, melodies
coming together to form the musical piece.
Two melodies sounded too thin (simple), and four or more melodies sounded less cohesive. 
% As mentioned earlier, there is no one right way to map the emotion profile of a novel to music. 
% Below we present the mapping rules we used to determine various parameters of the generated music.
% and the rationale behind them. 
% Not every parameter of the musical piece is determined by the emotional profile of the novel.
% We also present some ideas for future work pertaining to each of the
% mappings.

% \begin{itemize}
\subsection{Major and Minor Keys}
	Major keys generally create a more positive atmosphere in musical pieces, whereas minor
keys tend to produce pieces with more negative undertones 
\cite{Hunter10,Ali10,Gabrielsson01,Webster05}. 
% \cite{Hunter10,Hunter08,Ali10,Gabrielsson01,Webster05}. 
No consensus has been reached on whether
particular keys themselves (for example, A minor vs E minor) % , or G major vs F major) 
evoke different emotions, and if so, what emotions are evoked by which keys. 
For this
reason, the prototype of Transprose does not consider different keys; the chosen key for
the produced musical pieces is limited to either C major or C minor. (C major was chosen because it
is a popular choice when teaching people music. It is simple because it does not have any sharps. 
C minor was chosen as it is the minor counterpart of C major.)

Whether the piece is major or minor is determined by the ratio of the number of positive words to 
the number of negative words in the entire novel. If the ratio is higher than 1, C major is
used, that is, only pitches pertaining to C major are played. 
If the ratio is 1 or lower, C minor is used. %, that is only those pitches pertaining to c minor are played.

% Though this prototype does not use keys other than C major or C minor, 
Experimenting with keys other than C major and C minor is of interest for future work. Furthermore, the eventual intent is
to include mid-piece key changes for added effect. For example, changing the key from 
C major to A minor when the plot suddenly turns sad. The process of changing key
is called modulation. Certain transitions such as moving from C major to A minor
are commonly used and musically interesting.

\subsection{Melodies}
%	There are countless possibilities for deciding on the actual notes of the pieces. 
We use three melodies to capture the change in emotion word usage in the text. The notes in one melody are based on the overall
emotion word density (the emotion words associated with any of the eight emotions in the NRC Emotion Lexicon).
We will refer to this melody, which is intended to capture the overarching emotional movement,
as {\it melody o} or $M_o$ (the `o' stands for overall emotion).
The notes in the two other melodies, {\it melody e1} ($M_\text{\it e1}$) and {\it melody e2} ($M_\text{\it e2}$), 
are determined by the most prevalent and second most prevalent emotions in the text, respectively.
Precisely how the notes are determined is described in the next sub-section, but first we describe
how the octaves of the notes is determined.

The octave of melody o is proportional to the difference between the joy and sadness densities of the novel.
We will refer to this difference by {\it JS}.
% \begin{equation}
% \text{\it JS} = \text{\it JoyDensity}(\text{\it text}) - \text{\it SadDensity}(\text{\it text})\\
% \end{equation}
We calculated the lowest density difference ($\text{\it JS}_\text{\it min}$) and highest JS score ($\text{\it JS}_\text{\it max}$) in a collection of novels.
For a novel with density difference, {\it JS}, the
 score is linearly mapped to octave 4, 5, or 6 of a standard piano,
as per the formula shown below:
\begin{equation}
\text{\it Oct}(M_o) = 4 + r(\frac{(\text{\it JS} - \text{\it JS}_\text{\it min})*(6 - 4)}{\text{\it JS}_\text{\it max} - \text{\it JS}_\text{\it min}})
\end{equation}
\noindent The function $r$ rounds the expression to the closest integer.
Thus scores closer to $\text{\it JS}_\text{\it min}$ are mapped to octave 4, scores closer to $\text{\it JS}_\text{\it max}$
are mapped to octave 6, and those in the middle are mapped to octave 5.
% -0.006 and 0.008

The octave of $M_\text{\it e1}$ is calculated as follows:
\begin{equation}
\text{\it Oct}(M_\text{\it e1}) = 
\begin{cases}
    Oct(M_o) + 1,& \hspace*{-3mm} \text{\small if {\it e1} is joy or trust} \\
	Oct(M_o) - 1,& \hspace*{-3mm} \text{\small if {\it e1} is anger, fear,}\\ 
	& \hspace*{-3mm} \text{\small sadness, or disgust}\\
	Oct(M_o), & \hspace*{-3mm} \text{\small otherwise}
\end{cases}
\end{equation}
\noindent That is, $M_\text{\it e1}$ is set to:
\vspace*{-1mm}
\begin{itemize}
\item an octave higher than the octave of $M_o$ if $e_1$ is a positive emotion,
\vspace*{-2mm}
% Recall that higher octaves evoke a sense of positivity.
\item an octave lower than the octave of $M_o$ if $e_1$ is a negative emotion,
% Recall that lower octaves evoke a sense of negativity.
\vspace*{-2mm}
\item the same octave as that of $M_o$ if  $e_1$ is surprise or anticipation.
\end{itemize}
\vspace*{-1mm}
\noindent Recall that higher octaves evoke a sense of positivity, whereas lower octaves evoke a sense of negativity.
The octave of $M_\text{\it e2}$ is calculated exactly as that of $M_\text{\it e1}$, except that it is
based on the second most prevalent emotion (and not the most prevalent emotion) in the text.

\subsection{Structure and Notes}

% In this prototype, the ultimate goal was to pair emotionally active parts of the novel
% (parts where there were more emotions) with more active parts of the melody. The following
% choices were made with this goal in mind. 
% There are innumerable ways in which notes can be created and structured. 
% In the next three paragraphs, we provide a high-level overview
% of the structure of melodies created by TransProse. Additional details are provided immediately afterwards.

As mentioned earlier, TransProse generates three melodies that together make the musical piece for a novel.
The method for generating each melody is the same, with the exception that the three melodies ($M_o$, $M_\text{\it e1}$, and $M_\text{\it e2}$)
are based on the overall emotion density, predominant emotion's density, and second most dominant emotion's density, respectively.
We describe below the method common for each melody, and use {\it emotion word density} as a stand in for the appropriate density. 

Each melody
is made up of four sections, representing four sections of the novel (the beginning, early
middle, late middle, and end).%\footnote{Pieces with fewer and more sections in Section 9.} 
% However, 
% we found that using fewer than four sections produced greater variety in the notes of the melodies, but left the
%  listener unable to distinguish between different parts of the novel.
%  More sections, and thus more total sub-sections, created
%  less movement in the melodies, as the emotion densities being sampled were much
%  closer in value, often resulting in the sounding of the same note.}
In turn, each section is represented by four measures. Thus each measure corresponds to a quarter of a section (a sub-section).
% We will refer to these quarters as {\it sub-sections}.
A measure, as defined earlier, is a series of notes. The number of notes, the pitch of each note,
and the relative duration of each note are determined such that they reflect the emotion word densities in the corresponding part of the novel.

 {\it Number of Notes}:  
In our implementation, we decided to contain the possible note durations to whole notes, 
half notes, quarter notes, eighth notes, and sixteenth notes. A relatively high emotion density is represented 
by many notes, whereas a relatively low emotion density is represented by fewer notes. We first split the 
interval between the maximum and minimum emotion density for the novel into five equal parts 
(five being the number of note duration choices -- whole, half, quarter, eighth, or sixteenth). 
% % The number of notes in a measure is determined from the emotion density of the corresponding sub-section. 
% A relatively high emotion density is represented
%by many notes, whereas a relatively low emotion density is represented by fewer notes.
%In our implementation, 
%
%% lowest emotion densities
%% will result in measures of
%% only one whole-note, whereas the highest emotion densities
%% will result in measures of sixteen one-sixteenth-notes ($1/16^{\text{th}}$ notes). 
%we first split the interval between the maximum and minimum emotion density for the novel
%into five equal parts. 
Emotion densities that fall in the lowest interval are mapped to
a single whole note. Emotion densities in the next interval are mapped to two half-notes.
The next interval is mapped to four quarter-notes.
 And so on, until the densities in the last interval are mapped to sixteen sixteenth-notes ($1/16^{\text{th}}$).
The result is
shorter notes during periods of higher emotional activity (with shorter notes making the
piece sound more “active”), and longer notes during periods of lower emotional activity.

% The formula for this process is:
% Number of Notes Per Measure = Emotion Value of Measure mapped from Emotion Value Scale to the Number of Notes Scale
% y = (x - xmin)
% Where the Emotion Value Scale is from the lowest value to the highest value of that
% Emotion over the entire novel and the Number of Notes Scale is from one whole note to
% sixteen sixteenth notes (in order: 1 whole note, 2 half notes, 4 quarter notes, 8 eighth
% notes, or 16 sixteenth notes). 

% Recall that the emotion spline is a continuous function that captures the emotion word density throughout the novel.
% The spline can be split into sixteen parts corresponding to the sixteen section quarters of the novel.
% Since each measure represents a section quarter, the pitches of the notes in that measure are determined
% by the corresponding portion of the emotion spline.
% TransProse samples the corresponding portion of the emotion spline as many
% times as the number of notes in that measure. For example, if the measure is to have eight one-eight-notes, then the
% the emotion spline (in that corresponding section quarter) is sampled eight times. 

{\it Pitch}: If the number of notes for a measure is $n$, then the corresponding sub-section is partitioned into $n$ equal parts
and the pitch for each note is based on the emotion density of the corresponding sub-section.
Lower emotion densities are mapped to more consonant pitches in the key (C major or C minor),
whereas higher emotion densities are mapped to less consonant pitches in the same scale.
% As mentioned earlier, the emotion density is determined corresponding to each note by
% analyzing the corresponding sub-section in the novel.
% The lowest and highest values for the novel are mapped to the most consonant and least consonant
% notes in the scale. 
For example, if the melody is in the key of C major, then the lowest to highest emotion densities
are mapped linearly to the pitches C, G, E, A, D, F, B.
Thus, a low emotion value would
create a pitch that is more consonant and a high
emotion value would create a pitch that is more dissonant (more interesting and
unusual). 

% {\it Number of Sections}: Fewer than four sections produced greater variety in the notes of the melodies, but left the
% listener unable to distinguish between different parts of the novel.
% More sections created
% less movement in the melodies, as the points where emotions were being sampled were much
% closer together, often sampling a similar emotional count as the previous sampling
% position, and therefore sounding the same note.

{\it Repetition}: Once the four measures of a section are played, the same four measures are repeated
in order to create a more structured and melodic feeling. Without the
repetition, the piece sounds less cohesive. 
% Thus, each section has eight measures (four unique measures played twice),
% and the whole piece has $8 \times 4 = 32$ measures (sixteen unique measures played twice). 

% Since four unique measures are created for each section, each of the measures represents one-quarter of the section.
% We will refer to these quarters of a section as {\it sub-section}.
% Our goal is that the more prevalent an emotion is in a section of the novel, the more “active” the corresponding measures should sound. 
% Thus if the emotion density is high in a particular section quarter, then the corresponding measure
% is composed of a large number of notes, each played for a relatively short duration.
% If, on the other hand, the emotion density of the section quarter
% is low, then the corresponding measure is composed of a small number of notes,
% each played for a relatively longer duration.

% In order to
% create  some variety in rhythm and also to represent periods where the emotional levels
% were the same, adjacent notes of the same pitch and tempo were combined in half-measure
% groups. For example, if eight sixteenth notes (a half-measure) were all the same pitch,
% they would be combined into one note that was a half-measure (two combined whole notes)
% long. This half-measure decision was a subjective one; any shorter combination (i.e.
% combining four sixteenth notes of the same pitch) sounded too repetitive, and any longer
% combination (i.e. combining sixteen sixteenth notes of the same pitch) obscured the
% movement (or “activity”) that was happening at that moment.

\subsection{Tempo}
% The time signature of a piece indicates two things: a) how many beats are in a measure, and b) which note duration represents one beat. 
% The time signature is written as one number stacked on another number. The upper number is how many beats are per bar (or measure), 
% and the lower number is the note duration that represents one beat. For example, a time signature of 6/8 would mean there are six beats per measure, 
% and an eighth note represents one beat. One of the most common time signatures is 4/4, and it is referred to as {\it common time}.
We use a \setmetera{4}{4} time signature (common time) because it is one of the most popular time signatures.
Thus each measure (sub-section) has 4 beats. We determined tempo (beats per minute)
by first determining how active the target novel is.
Each of the eight basic emotions is assigned to be either active, passive, or neutral. 
In TransProse, the tempo
% , or the speed of the piece, 
is proportional to the {\it activity score}, which we define to be the difference between the average density of the active
emotions (anger and joy) and the average density of the passive emotions (sadness).
% We will refer to this score as the {\it activity score}.
The other five emotions (anticipation, disgust, fear, surprise, and trust) were
considered ambiguous or neutral, and did not influence the tempo.  

% TransProse generates melodies with relatively high tempos for novels with 
% relatively high activity scores. 
We subjectively identified upper and lower bounds for
the possible tempo values to be 180 and 40 beats/minute, respectively. 
% Thus we linearly mapped activity scores between -0.002 and 0.017 to 40 and 180 beats/minute.
We determined activity scores for a collection of novels, and identified the highest activity score
($\text{\it Act}_\text{\it max}$) and the lowest activity score ($\text{\it Act}_\text{\it min}$).
For a novel whose activity score was $\text{\it Act}$, we determined tempo as per the formula shown below:
\begin{equation}
% tempo = 40 + \frac{(x - (-0.002))*(180 - 40)}{0.017 - (-0.002)}
\text{\it tempo} = 40 + \frac{(\text{\it Act} - \text{\it Act}_\text{\it min})*(180 - 40)}{\text{\it Act}_\text{\it max} - \text{\it Act}_\text{\it min}}
\end{equation}
\noindent 
Thus, high activity scores were represented by tempo values closer to 180
and lower activity scores were represented by tempo values closer to 40.
The lowest activity score in our collection of texts, $\text{\it Act}_\text{\it min}$, was -0.002 whereas the highest activity score, $\text{\it Act}_\text{\it max}$, was 0.017.
% Eventually, a corpus
% rating the “activity” of each word will influence the tempo as well.

% For prototyping purposes, the pieces are all currently in the 4/4 time signature,
% which is a standard time signature; additional signatures will be explored in future
% iterations.

\section{Converting Specifications to Music}
JFugue is an open-source Java API that helps create generative music.\footnote{\url{http://www.jfugue.org}}
It allows the user to easily experiment with 
different notes, instruments, octaves, note durations, etc within a Java program. %, and then exports the finished piece to a MIDI file.
 JFugue requires a line of specifically-formatted text that describes the melodies in order to play them. 
The initial portion of the string of JFugue tokens for the novel {\it Peter Pan} is shown below. The string conveys the 
overall information of the piece as well as the first eight measures (or one section) for each of the three
melodies (or voices). 
\begin{quote}
KCmaj X[VOLUME]=16383 V0 T180 A6/0.25 D6/0.125 F6/0.25 B6/0.25 B6/0.125 B6/0.25 B6/0.25...
\end{quote}
\noindent $K$ stands for key and {\it Cmaj} stands for C major. This indicates that the rest
of the piece will be in the key of C major. The second token controls the volume, which in
this example is at the loudest value (16383). {\it V0} stands for the first melody (or voice). 
The tokens with the letter $T$ indicate the tempo, which in the case of
this example is 180 beats per minute. 

The tokens that follow indicate the notes of the melody. The letter is
the pitch class of the note, and the number immediately following it is the octave. The
number following the slash character indicates the duration of the note. (0.125 is an
eighth-note (1/8th), 0.25 is a quarter note, 0.5 is a half note, and 1.0 is a whole note.) 
% These durations add up to 4 whole notes (four measures) and then the pattern repeats, making
% eight measures, or a section. The pitches for the second and third voices (V1 and V2) then
% follow in the same manner.
We used JFugue to convert the specifications of the melodies into music.  
JFugue saves the pieces as a midi files, which we converted to MP3 format.\footnote{The MP3 format
uses a lossy data compression, but the resulting files are significantly smaller in size. Further,
a wider array of music players support the MP3 format.}
\iffalse
% \iftrue
[11] "F6/0.125"
[12] "D6/0.125"
[13] "A6/0.125"
[14] "E6/0.125"
[15] "E6/1.0"
[16] "D6/0.25"
[17] "A6/0.5"
[18] "E6/0.25"
[19] "V0"
[20] "T210"
[21] "A6/0.25"
[22] "D6/0.125"
[23] "F6/0.25"
[24] "B6/0.25"
[25] "B6/0.125"
[26] "B6/0.25"
[27] "B6/0.25"
[28] "F6/0.125"
[29] "D6/0.125"
[30] "A6/0.125"
[31] "E6/0.125"
[32] "E6/1.0"
[33] "D6/0.25"
[34] "A6/0.5"
[35] "E6/0.25"
[36] "V1"
[37] "T210"
[38] "D7/0.25"
[39] "F7/0.25"
[40] "B7/0.25"
[41] "B7/0.25"
[42] "B7/0.25"
[43] "B7/0.125"
[44] "F7/0.125"
[45] "D7/0.125"
[46] "A7/0.125"
[47] "E7/0.125"
[48] "G7/0.125"
[49] "C7/1.0"
[50] "C7/0.5"
[51] "G7/0.5"
\fi

% \iffalse
 \iftrue
\begin{table*}[t!]
\caption{Emotion and audio features of a few popular novels that were processed by TransProse.
%  Open pdf in the freely available Adobe Reader to access the audio embeddings in the last column, {\it Piece}. Alternatively, all of
The musical pieces are available at: \url{http://transprose.weebly.com/final-pieces.html}.}
\label{tab:pieces}
\begin{center}
% \small{
\resizebox{0.95\textwidth}{!}{
%\vspace*{-2mm}
\begin{tabular}{lcccc cccc c}
\hline
%     {\bf Novel}   &{\bf Piece} &{\bf Negative} &{\bf Neutral} &{\bf Total}\\
% To Kill A Mockingbird & \attachfile{To\ Kill\ A\ Mockingbird.mp3}\\
% To Kill A Mockingbird & \attachfile{ToKill.mp3}\\

Book Title	&Emotion 1	&Emotion 2	&Octave	&Tempo	&Pos/Neg	&Key	&Activity &Joy-Sad \\
\hline
{\it A Clockwork Orange}	&Fear	&Sadness	&5	&171	&Negative 	&C Minor	&0.009	&-0.0007 \\
% {\it Adventures of Sherlock}	&Trust	&Fear	&5	&167	&Positive	&C Major	&0.009	&0.0008 &\attachfile{mp3s/Sherlock.mp3}\\
% \;\;\; {\it Homes, The} & & & & & & & & &\\
{\it Alice in Wonderland}	&Trust	&Fear	&5	&150	&Positive	&C Major	&0.007	&-0.0002 \\
{\it Anne of Green Gables}	&Joy	&Trust	&6	&180	&Positive	&C Major	&0.010	&0.0080 \\
% {\it Einstein's Dreams}	&Trust	&Joy	&6	&180	&Positive	&C Major	&0.010	&0.0066 &\attachfile{mp3s/Einstein.mp3}\\
{\it Heart of Darkness}	&Fear	&Sadness	&4	&122	&Negative	&C Minor	&0.005	&-0.0060 \\
{\it Little Prince, The}	&Trust	&Joy	&5	&133	&Positive	&C Major	&0.006	&0.0028 \\
{\it Lord of The Flies}	&Fear	&Sadness	&4	&151	&Negative	&C Minor	&0.008	&-0.0053 \\
% {\it Norwegian Wood}	&Trust	&Joy	&5	&172	&Positive	&C Major	&0.009	&0.0030 &\attachfile{mp3s/Norwegian.mp3}\\
% Old Man and the &Trust	&Fear	&5	&134	&Negative	&C Minor	&0.006	&-0.0011 &\attachfile{mp3s/oldman.mp3}\\
% Sea, The & & & & & & & & &\\
{\it Peter Pan}	&Trust	&Joy	&6	&180	&Positive	&C Major	&0.010	&0.0040 \\
{\it Road, The}	&Sadness	&Fear	&4	&42	&Negative	&C Minor	&-0.002	&-0.0080 \\
{\it To Kill a Mockingbird}	&Trust	&Fear	&5	&132	&Positive	&C Major	&0.006	&-0.0013 \\
\hline
\end{tabular}
}
\end{center}
 \vspace*{-3mm}
\end{table*}
\fi

\section{Case Studies}
%    -- case studies on specific novels
%    	-- embed music
%	-- show visualization of emotions
%	-- what does the output of transprose for each piece of text show?
%	-- select case studies to show differences in how happy, sad, angry, trust, etc pieces of text are captured by TransProse.
%    	-- can we ask unbiased people questions about the output to test whether it is functioning suitably. 
%		-- For example, we can show them two version of Transprose on the same piece of text by varying something in the algorithm, and asking which output is better (or something more specific)?'

We created musical pieces for several popular novels through TransProse.
These pieces are available at: http://transprose.weebly.com/final-pieces.html.
Since these novels are likely to have been read by many people, 
the readers can compare their understanding of the story with the music generated by TransProse.
Table \ref{tab:pieces} presents details of some of these novels.
% Some of generated musical pieces are embedded in this pdf in Table \ref{tab:pieces}.
% Double click the icons to listen. 

% (Any audio player such as iTunes or QuickTime that supports MP3 format can play the music.)
% Note: The audio embedding may not be supported by pdf viewers other than
% the freely available Adobe Reader.\footnote{http://get.adobe.com/reader} Alternatively, all of
% these pieces are also available at this url: http://transprose.weebly.com/final-pieces.html.

\subsection{Overall Tone}

	TransProse captures the overall positive or negative tone of the novel by assigning an
either major or minor key to the piece. {\it Peter Pan} and {\it Anne of Green Gables}, 
novels with overall happy and uplifting moods, created pieces in the major key. On the
other hand, novels such as {\it Heart of Darkness, A Clockwork Orange}, and {\it The Road}, with dark
themes, created pieces in the minor key. The effect of this is pieces that from the start
have a mood that aligns with the basic mood of the novel they are based on.

\subsection{Overall Happiness and Sadness}
	
	The densities of happiness and sadness in a novel are represented in the baseline
octave of a piece. This representation instantly conveys whether the novel has a
markedly happy or sad mood. The overall high
happiness densities in {\it Peter Pan} and {\it Anne of Green Gables} create pieces in an
octave above the average, resulting in higher tones and a lighter overall mood. Similarly,
the overall high sadness densities in {\it The Road} and {\it Heart of Darkness} result in pieces an
octave lower than the average, and a darker overall tone to the music. Novels, such
as {\it A Clockwork Orange}, and {\it The Little Prince}, where
happiness and sadness are not dramatically higher or lower than the average novel
remain at the average
octave, allowing for the creation of a more nuanced piece.

\subsection{Activeness of the Novel}
	
%	The general pace of the novel was captured by the tempo. 
Novels with lots of active emotion words, such as
	{\it Peter Pan, Anne of Green
Gables, Lord of the Flies,} and {\it A Clockwork Orange}, generate fast-paced pieces with tempos over 170 beats per minute. On
the other hand, {\it The Road}, which has relatively few active emotion words is rather slow (a tempo of 42
beats per minute).

\subsection{Primary Emotions}
	
%	A significant part of each piece is influenced by the novel’s top two emotions. Each
The top two emotions of a novel inform two of the three melodies in a piece ($M_\text{\it e1}$ and $M_\text{\it e2}$). 
Recall that 
% the octave of
% a melody is determined from the emotion it is based on. 
if the melody is based on a positive
 emotion, it will be an octave higher than the octave of $M_o$, and if it is based on a negative emotion, it will be
 an octave lower. 
For novels where the top two emotions are both positive, such as
{\it Anne of Green Gables} (trust and joy), the pieces sound especially light and joyful. For
novels where the top two emotions are both negative, such as {\it The Road} (sadness and fear),
the pieces sound especially dark.

\subsection{Emotional Activity}

	Unlike the overall pace of the novel, individual segments of activity were also
identified in the pieces through the number and duration of notes (with more and shorter
notes indicating higher emotion densities). 
This can be especially heard in the third section of {\it A Clockwork Orange}, the final section
of {\it The Adventures of Sherlock Holmes}, the second section of {\it To Kill a Mockingbird}, and the
final section of {\it Lord of the Flies}.
In {\it A Clockwork Orange}, the main portion of the
piece is chaotic and eventful, likely as the main characters cause havoc; at the end of
the novel (as the main character undergoes therapy) the piece dramatically changes and
becomes structured. Similarly, in {\it Heart of Darkness}, the piece starts out only playing a
few notes; as the tension in the novel builds, the number of notes increases and their durations
decrease.

\section{Comparing Alternative Choices}
% \section{Investigating Alternative Choices  in the Mapping of Text to Music}
We examine choices made in TransProse by comparing musical pieces generated with
different alternatives. These audio clips are available here:
 \url{http://transprose.weebly.com/clips.html}.
% Size constraints prevented the embedding of these additional audio clips in this pdf.

% TransProse generates pieces with three melodies.  
Pieces with two melodies (based on overall emotion density and the predominant emotion's density)
and pieces based on four melodies (based on the top three emotions and the overall emotion density) were 
generated and uploaded in the clips webpage.
% Here, they can be compared with the official TransProse output that uses three melodies.
Observe that with only two melodies, the pieces tend to sound thin, whereas
with four melodies the pieces sound less cohesive and sometimes chaotic.
% ({\it Heart of Darknesses} and {\it Alice in Wonderland} and {\it A Clockwork Orange - 4 melodies}.
 The effect of increasing and decreasing the total number of sections and sub-sections is also presented.
%  through pieces based on eight sections and pieces based on two sections.
% We found pieces with eight sections created less melodic movement,
% and pieces with only two sections did not distinguish well between different parts of the novel.
% (see “Heart of Darkness - 2 sections” and “A Clockwork Orange - 2 sections”).
Additionally, the webpage displays pieces with tempos and octaves beyond
the limits chosen in TransProse. We also show other variations such as pieces for relatively positive novels generated in C minor (instead of C major).
These alternatives are not necessarily incorrect, but they tend to often be less effective.
% and sometimes they can be used with good effect. 
% However, direct comparisons allows us to see that some choices are clearly
% better than others.

% Our goal, with TransProse, was to present choices that have ked well.
% -Lowering the bottom end of the octave scale created pieces that were too low to be enjoyable (see “The Road - Lower Bottom Octave Bound”).
% -Raising the top end of the octave scale created pieces that were too high to be enjoyable (see “Anne of Green Gables - Higher Upper Octave Bound”).

\section{Limitations and Future work}
We presented a system, TransProse, that generates music according to the
use of emotion words in a given piece of text.
% Currently, TransProse produces piano music with three piano melodies
% that combine to form the musical piece. Even though there is no single correct
% method to generate music from text, there are numerous possibilities. TransProse
% uses known relations between parameters of music and the emotions they evoke,
% such as using major keys to express positive sentiment, higher tempo to 
% express activity, and using consonant and disconsonant notes to capture
% the change in distribution of emotion words in the input text.
A number of avenues for future work exist such as exploring the use of
mid-piece key changes and intentional harmony and discord between the melodies.
% of musical instruments other than the piano, adding more complexity to the generated
% musical piece by varying major and minor keys, and so on.
We will further explore ways to capture activity in music. For example, an automatically 
generated activity lexicon (built using the method proposed by Turney and Littman \shortcite{TurneyL03}) can be used to identify
portions of text where the characters are relatively active (fighting, dancing, conspiring, etc) and 
areas where they are relatively passive (calm, incapacitated, sad, etc).
One can even capture non-emotional features of the text in music. For example, % a sudden burst
% of conversation with short utterances can be captured with higher tempo and more discordant notes.
recurring characters or locations in a novel could be indicated by recurring motifs.
% Also, semantic content of the novel can be used to generate appropriate music---for example, 
% playing music with distinctive characteristics 
% when a particular chapter is about church and God or when it is about
% gallant heroes.
% The voices presented in this paper are largely independent of each other,
% but exploring the interaction between the voices is interesting future work.
We will conduct human evaluations asking people to judge
various aspects of the generated music such as the quality of music and the amount and type of emotion 
evoked by the music. We will also evaluate the impact of textual features such as the length of the novel
and the style of writing on the generated music. 
Work on capturing note models (analogous to language models) from existing pieces of music and using them to improve the music generated by TransProse seems especially promising.

 \bibliographystyle{acl}
 \bibliography{references}

\end{document}